\documentclass[runningheads]{llncs}
\usepackage{amsmath}
\usepackage[T1]{fontenc}
\usepackage{pgfplots}
\usepgfplotslibrary{groupplots}
\pgfplotsset{compat=1.18}
\usepackage{cite}

\usepackage{xcolor}
\definecolor{mblue}{HTML}{4A7FB5}
\definecolor{morange}{HTML}{E08214}
\definecolor{mgreen}{HTML}{1B7837}

\usepackage{graphicx,verbatim}
\usepackage{multirow}
\usepackage{amssymb}
\usepackage{booktabs}

\usepackage{xcolor}
\usepackage{hyperref}
\usepackage{orcidlink}

\hypersetup{
    colorlinks=true,
    linkcolor=blue,   
    citecolor=blue,   
    urlcolor=blue      
}

\usepackage{marvosym}

\begin{document}
\title{X-LMC: Cross-View Spatiotemporal Collateral Circulation Scoring from DSA}

\titlerunning{X-LMC: Cross-View Spatiotemporal Collateral Scoring from DSA}

%

\author{Maedeh Hafezi Moghadas\inst{1}\textsuperscript{(\Letter)}\orcidlink{0009-0004-9806-4391} \and
Hakim Baazaoui\inst{2}\orcidlink{0000-0002-6820-2659} \and
Lukas Bastian Otto\inst{2}\orcidlink{0009-0008-9598-0561} \and
Susanne Wegener\inst{2}\orcidlink{0000-0003-4369-7023} \and
Björn Menze\inst{3}\orcidlink{0000-0003-4136-5690}
\and
Ezequiel De la Rosa\inst{3}\orcidlink{0000-0002-9042-1962} }


%
\authorrunning{M. Hafezi Moghadas et al.}
%
\institute{Friedrich-Alexander-Universität, Erlangen-Nürnberg, Germany
\email{maedeh.hafezi@fau.de}
\and
University Hospital Zurich, Zurich, Switzerland
\\
\and
University of Zurich, Zurich, Switzerland
}

\begingroup
\renewcommand{\thefootnote}{}
\footnotetext{M. Hafezi Moghadas and H. Baazaoui---Equal contribution.}
\footnotetext{B. Menze and E. De la Rosa---Shared senior authorship.}
\endgroup

\maketitle              
\begin{abstract}

Digital subtraction angiography (DSA) is the reference standard for leptomeningeal collateral (LMC) assessment, providing critical prognostic insights to guide secondary treatment strategies, neurorehabilitation planning, and retrospective stroke research. However, clinical LMC grading via the ASITN/SIR scale relies on manual, highly variable visual inspection. We introduce X-LMC, a spatiotemporal framework for automated collateral scoring from time-resolved biplane DSA. The proposed architecture encodes spatial frame representations through a DINOv2 backbone, fuses orthogonal projections via a token-level cross-view attention module, and models representations of contrast bolus dynamics using a recurrent network architecture.  We evaluate our framework on a multicenter dataset of 134 patients with M1-segment occlusions. In a 5-fold cross-validation setting, X-LMC yields higher point estimates than static architectures and spatiotemporal baselines adapted from related angiographic tasks, achieving a Quadratic Weighted Kappa (QWK) of 0.398 (vs. 0.322) and a dichotomized macro-F1 score of 0.711 (vs. 0.663) against the best-performing baseline. X-LMC performance also aligns with the observed clinical inter-rater agreement (QWK: 0.314). As the first DSA study attempting to automate LMC scoring, we demonstrate that multi-view temporal deep learning can capture collateral-specific contrast kinetics. Ultimately, these benchmarks delineate the clinical ambiguities and achievable performance boundaries of automated ASITN/SIR grading, establishing a reproducible foundation for objective hemodynamic phenotyping in stroke cohorts. Code is available at \url{https://github.com/maedehafezi/X-LMC}.


\end{abstract}

\section{Introduction}
Acute ischemic stroke (AIS) remains a leading cause of mortality and disability worldwide~\cite{gbd2021global}. Therapeutic approaches focus on rapid recanalization of an occluded blood vessel through endovascular treatment or intravenous thrombolysis. The viability of brain tissue critically depends on collateral flow, and especially on leptomeningeal collaterals (LMCs) at the pial surface of the brain that are ``recruited'' following an arterial obstruction. They expand due to pressure gradient changes and provide retrograde blood flow, thereby delaying the progression from penumbra to irreversible infarction~\cite{liebeskind2003collateral}. Despite their prognostic significance, LMCs fall below the spatial resolution of conventional neuroimaging due to their small caliber ($<1$ mm) and pial location. Consequently, clinicians must rely on indirect surrogates for LMC assessment, such as retrograde filling in computed tomography angiography (CTA) or perfusion delay maps \cite{menon2011regional,lin2020perfusion}.

Intra-procedural digital subtraction angiography (DSA) is the reference standard for evaluating LMCs~\cite{liu2018guidelines}. It provides high spatial resolution ($0.08 \times 0.08$ to $0.2 \times 0.2 \text{ mm}^2$ pixels), enabling precise visualization of the collateral microvasculature. Beyond supporting stroke etiology identification~\cite{shi2022quantitative}, collateral status supports secondary treatment and rehabilitation decision making. However, clinical grading via the reference American Society of Interventional and Therapeutic Neuroradiology/Society of Interventional Radiology (ASITN/SIR) scale~\cite{higashida2003trial} remains manual, time-intensive, and dependent on specialized expertise.
Prior work reported poor interobserver agreement for angiographic ASITN/SIR collateral assessment, with $\kappa = 0.16 \pm 0.0065$ overall and $\kappa = 0.27 \pm 0.014$ after dichotomization into poor versus good collaterals~\cite{hassen2019inter}, limiting reproducible large-scale cohort analysis. Automated collateral scoring has been mostly studied in (single/multi-phase) CTA and CT perfusion ~\cite{su2020automatic,potreck2022rapid}. However, these modalities provide \emph{indirect} LMC surrogates, with limited spatial resolution compared to DSA.  
In DSA, deep learning has been applied to diverse tasks, including artery–vein segmentation~\cite{su2024cave,liu2024dias} and thrombus detection~\cite{mittmann2022deep}. For reperfusion grading, frameworks automate (modified) Thrombolysis in Cerebral Infarction ((m)TICI) scoring via spatiotemporal recurrent networks (\textit{DeepTICI})~\cite{nielsen2021deep}, leveraging intensity-projection perfusion maps~\cite{su2021autoTICI}, or using cross-view fusion modules in \textit{CVFSNet}~\cite{xu2025cvfsnet}. While graph neural networks were recently introduced to segment collateral branches~\cite{cao2026leptomeningeal}, direct grading of LMCs in DSA remains unaddressed.

Herein, we introduce X-LMC, a cross-view spatiotemporal framework for automated pre-interventional ASITN/SIR collateral scoring from biplane DSA series. By incorporating a bidirectional cross-view attention (X-Attn) module, the architecture dynamically exchanges token-level information between synchronized anteroposterior (AP) and lateral (Lat) projections, capturing complex contrast bolus kinetics within the microvasculature. As a first proof-of-concept for automated collateral assessment on standard DSA, this work establishes a reproducible foundation for large-scale hemodynamic phenotyping. Our main contributions are threefold: (i)~the first deep learning framework to derive ordinal ASITN/SIR collateral scores directly from raw 2D+time biplane DSA series; (ii)~a multi-view architecture employing a DINOv2~\cite{oquab2023dinov2} vision backbone with bidirectional X-Attn modules to integrate orthogonal angiographic projections prior to spatiotemporal modeling; and (iii)~a comprehensive evaluation against static and spatiotemporal baselines, benchmarked directly against human inter-rater agreement in this clinically complex task.

\section{Methods \& Experimental Setup}

We formulate the LMC scoring task as a multi-view spatiotemporal learning problem. Let $\mathcal{X} \subset \mathbb{R}^{T \times H \times W \times C}$ denote the space of time-resolved DSA sequences, where $T$, $H$, $W$, and $C$ denote the number of temporal frames, spatial height, spatial width, and input channels, respectively. For each patient, we consider a paired observation $\mathbf{I} = \{I_{\mathrm{AP}}, I_{\mathrm{Lat}}\} \in \mathcal{X}^2$, corresponding to simultaneous biplane projections in the AP and Lat planes, providing orthogonal perspectives of the cerebral vasculature. Our objective is to learn a mapping $f_\theta: \mathcal{X}^2 \rightarrow \mathcal{Y}$, parameterized by weights $\theta$, mapping angiographic dynamics to a discrete collateral score $y \in \mathcal{Y}$. The target labels follow the ASITN/SIR scale, the most widely adopted clinical scale for grading collateral flow in DSA based on the completeness and speed of retrograde contrast arrival to the ischemic territory~\cite{higashida2003trial}, treated as an ordered label space $\mathcal{Y} = \{0, 1, \dots, 4\}$, where $y=0$ represents the absence of collateral flow and $y=4$ denotes complete and rapid contrast arrival. The proposed framework $f_\theta$ jointly optimizes spatial feature extraction from multi-view inputs and the modeling of contrast bolus kinetics over time, ensuring that the predicted score $\hat{y}$ captures both the anatomical extent of the collateral network and the temporal blood flow dynamics within the ischemic territory.

\subsection{Spatiotemporal Framework}
\begin{figure}
\centering
\includegraphics[width=12.2 cm]{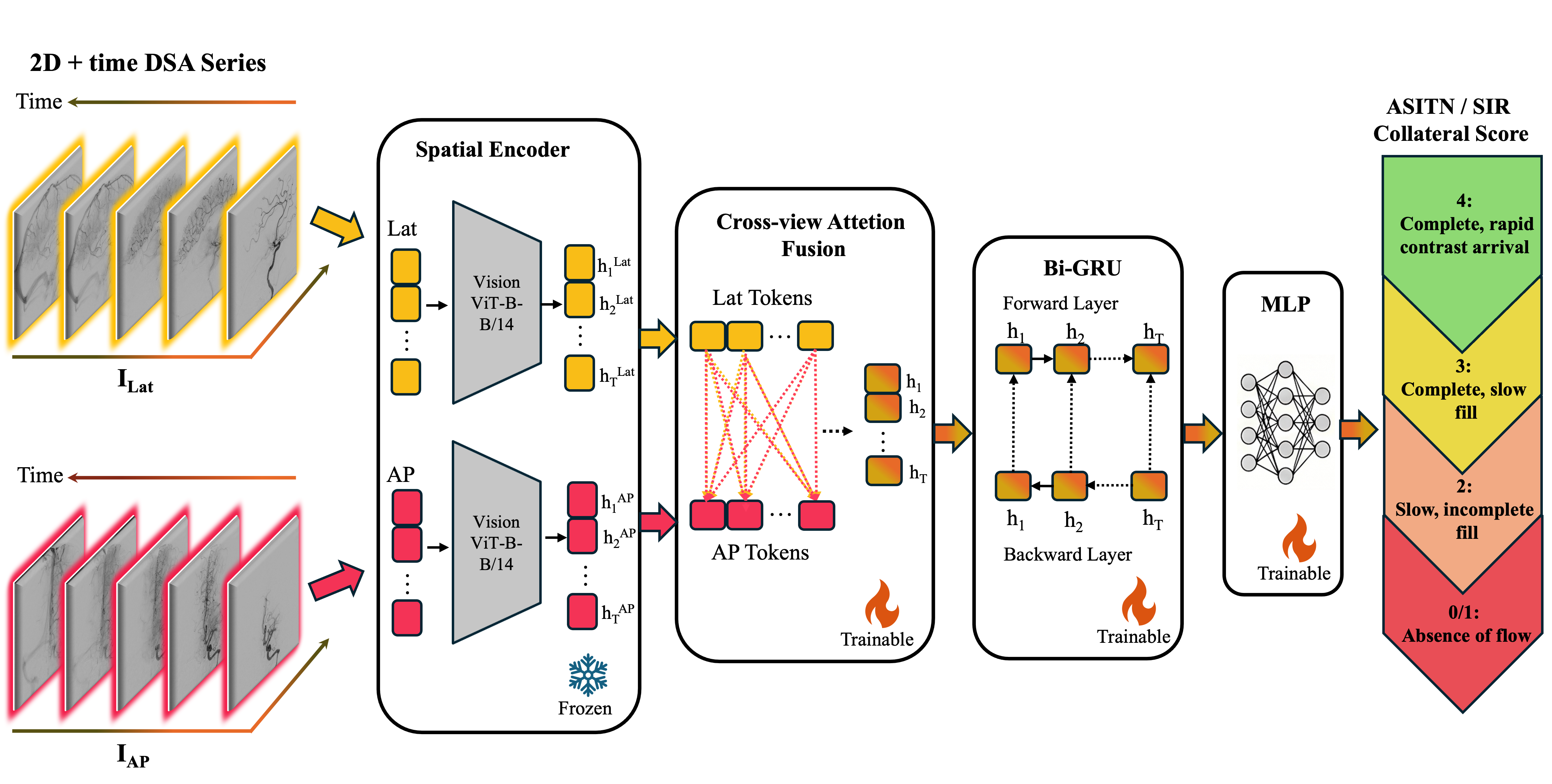}
\caption{Overview of the proposed X-LMC framework for collateral scoring.}
\label{fig:spatiotemporal_arch}
\end{figure}

The mapping $f_\theta$ is implemented as an end-to-end framework designed to process the paired biplane observation $\mathbf{I} = \{I_{\mathrm{AP}}, I_{\mathrm{Lat}}\}$ of variable temporal length $T$. The architecture consists of two symmetric branches for parallel feature extraction from the sequences in $\mathcal{X}$, followed by a temporal reasoning module operating on the derived cross-view fused representation. The pipeline proceeds in three stages: (i) frame-wise spatial encoding, where features are extracted independently from each view $I_v$; (ii) per-timepoint cross-view feature interaction, where information is exchanged between orthogonal projections to produce view-enhanced feature representations, which are subsequently aggregated into a unified representation $h_t$; and (iii) bidirectional temporal modeling, to capture the temporal contrast-flow dynamics across the sequence length $T$ (Fig.~\ref{fig:spatiotemporal_arch}).

\subsubsection{Multi-view spatial encoding.}To obtain robust descriptors of the cerebral vasculature, we utilize DINOv2, a self-supervised Vision Transformer (ViT-B/14), as the spatial encoder for each angiographic view. Each frame $f_t^{(v)} \in I_v$ is processed independently to extract token-level representations. Specifically, the frame is partitioned into non-overlapping patches and prepended with a learnable class token, which, after processing through the transformer layers, yields a sequence of latent token embeddings $Z_t^{(v)} \in \mathbb{R}^{N \times D}$, where $N$ denotes the number of tokens including the class token and $D=768$ is the embedding dimension. These token-level representations are retained for subsequent cross-view feature interaction, rather than reducing each frame to a single global descriptor before fusion. Due to the limited size of our cohort and the high capacity of the ViT-B/14 backbone, the encoder weights are kept frozen to serve as a fixed feature extractor, thereby preventing overfitting to the training set.

\subsubsection{Feature fusion.}To integrate complementary information from orthogonal angiographic projections while preserving temporal alignment, we employ bidirectional cross-view attention (X-Attn)~\cite{vaswani2017attention}. For each timepoint $t \in \{1,\ldots,T\}$, X-Attn is applied at the token level, allowing AP and Lat embeddings to mutually attend to one another. Given query $Q = Z_q W_Q$, key $K = Z_{kv} W_K$, and value $V = Z_{kv} W_V$ projections with learnable matrices $W_Q, W_K, W_V$, the attention output is computed as $\mathrm{Attention}(Q,K,V) = \mathrm{softmax}\left(QK^\top / \sqrt{d_k}\right)V$, and is applied bidirectionally as $\widetilde{Z}_t^{(\mathrm{AP})}=\mathrm{X\text{-}Attn}(Z_t^{(\mathrm{AP})},Z_t^{(\mathrm{Lat})})$ and $\widetilde{Z}_t^{(\mathrm{Lat})}=\mathrm{X\text{-}Attn}(Z_t^{(\mathrm{Lat})},Z_t^{(\mathrm{AP})})$. The updated class-token embeddings are summed to obtain the fused frame-level representation $h_t = \widetilde{z}_{t,0}^{(\mathrm{AP})} + \widetilde{z}_{t,0}^{(\mathrm{Lat})} \in \mathbb{R}^{D}$. The resulting sequence $\mathcal{H} = \{h_1,h_2,\ldots,h_T\}$ serves directly as input to the temporal modeling module.

\subsubsection{Perfusion dynamics via temporal learning.}To model the temporal evolution of contrast dynamics across the sequence length $T$, we process the fused frame-level representations $\mathcal{H}$ using a bidirectional gated recurrent unit (Bi-GRU)~\cite{cho2014learning}, integrating contextual information from both earlier and later time\-steps to capture the overall hemodynamic progression of contrast flow. For a GRU with hidden dimension $H_d$, the final forward ($\overrightarrow{h}_T$) and backward ($\overleftarrow{h}_1$) hidden states are concatenated to form the global spatiotemporal representation $h_{\mathrm{seq}} = [\overrightarrow{h}_T \parallel \overleftarrow{h}_1] \in \mathbb{R}^{2H_d}$. This representation is passed to an MLP classifier for collateral grade prediction.

\subsection{Dataset and preprocessing} Due to the scarcity of public DSA datasets with collateral grading, we used data from the multicenter acute stroke MAGIC repository~\cite{baazaoui2025multicentre}, including anonymized data of AIS patients. We included pre-interventional DSA imaging of 134 patients undergoing endovascular treatment for AIS. Each case consists of simultaneous biplane DSA series acquired via a single contrast injection. Collateral circulation was graded using the ASITN/SIR scale (0--4) by a fellow in interventional neuroradiology and a neurology resident specializing in stroke. For supervised training and evaluation, the score assigned by the interventional neuroradiology fellow was used as the reference label. A subset of cases (N=98) was independently graded by both raters and used to quantify inter-rater agreement.

To minimize variability, we restricted the dataset to M1-segment middle cerebral artery (MCA) occlusions. 
To harmonize acquisitions, series were temporally standardized to 2 frames/s via linear interpolation and truncated at 30 frames. Pre-contrast baseline frames were removed to prioritize arterial and early venous phases, preserving hemodynamics critical for collateral assessment. Finally, images were resized to $224 \times 224$ pixels and normalized to $[0, 1]$ for compatibility with standard deep learning backbones.

\subsection{Experiments and baselines}

We evaluate the model on two tasks: multiclass ASITN/SIR grading (0--4) and binary classification of \emph{good} ($n=66$) versus \emph{poor} (ASITN/SIR $\leq 2$, $n=68$) collateralization. Due to its low prevalence (2.24\%), grade 0 was merged with grade 1. Model performance was contextualized with inter-rater agreement and compared against: (i) a mean-guess baseline; and (ii) an adapted TICI-style spatiotemporal baseline inspired by Nielsen et al.~\cite{nielsen2021deep}. This baseline adopts the high-level DeepTICI-inspired architecture, consisting of dual-branch frame-wise feature extraction followed by recurrent temporal modeling. Specifically, it uses $1 \times 1$ convolutional fusion and a unidirectional GRU (Uni-GRU), but is adapted without DeepTICI’s frame-wise pseudo-perfusion supervision or TICI-specific loss formulation. All models were evaluated via 5-fold cross-validation, using four folds for training/validation and one fold for independent testing.

\subsubsection{Metrics.}
Multiclass performance is evaluated using mean absolute error (MAE) and accuracy within a $\pm1$ grade tolerance ($\mathrm{ACC}_{\pm1}$)~\cite{nielsen2021deep}, reflecting clinical acceptability of minor discrepancies. To account for the ordinal nature of the ASITN/SIR scale, we report quadratic weighted kappa (QWK) with weights $w_{ij} = (i-j)^2 / (N-1)^2$. For the binary task (\emph{good} vs. \emph{poor}), we report macro-averaged F1-score (mF1) and Cohen's kappa (Kappa). Both QWK and Kappa are calculated as $1 - \sum w_{ij}O_{ij} / \sum w_{ij}E_{ij}$, where $O_{ij}$ and $E_{ij}$ are the observed and expected agreements, respectively.

\subsubsection{Implementation.}
We utilized a DINOv2 ViT-B/14 backbone as the spatial feature extractor of our model, initialized with pretrained weights and kept frozen during training. Temporal dependencies were modeled using a single-layer Bi-GRU with 256 hidden units. Data augmentation included horizontal flipping, random shifts, scaling, rotation, and contrast perturbations. Optimization was performed using Adam with a batch size of 1. The model was trained using a hybrid loss combining cross-entropy with an auxiliary MAE term on the predicted class probabilities,
$\mathcal{L} = (1-\alpha)\mathcal{L}_{\mathrm{CE}} + \alpha\mathcal{L}_{\mathrm{MAE}}$,
with $\alpha = 0.3$ selected empirically on the validation set.
Models were trained for a maximum of 1000 epochs with early stopping on an NVIDIA Tesla V100 SXM2 (32\,GB) GPU.

\section{Results}

\begin{figure*}[t]
\centering

\begin{tikzpicture} 
\begin{groupplot}[
    group style={
        group size=3 by 1,
        horizontal sep=2cm,
    },
    width=0.295\textwidth,
    height=3.6cm,
    ymin=0.3, ymax=4.7,
    ytick={1,2,3,4},
    yticklabels={X-LMC(ours), TICI-style, Inter-rater, Mean guess},
    yticklabel style={font=\fontsize{8pt}{\baselineskip}\selectfont},
    xticklabel style={font=\fontsize{8pt}{\baselineskip}\selectfont},
    y dir=normal,
    tick label style={font=\fontsize{8pt}{\baselineskip}\selectfont},
    grid=major,
    grid style={gray!15},
    axis line style={gray!60},
    error bars/x dir=both,
    error bars/x explicit,
    clip=false  
]
\nextgroupplot[
    title={(a) QWK},
    xmin=-0.28, xmax=0.68,
]
\addplot[draw=none, fill=yellow!12, forget plot] coordinates {
    (0.21,0.3) (0.40,0.3) (0.40,4.7) (0.21,4.7)} -- cycle;
\addplot[draw=none, fill=orange!10, forget plot] coordinates {
    (0.41,0.3) (0.60,0.3) (0.60,4.7) (0.41,4.7)} -- cycle;
\addplot[black!50, dashed, forget plot] coordinates {(0,0.3) (0,4.7)};
\addplot+[gray, mark=*, mark options={draw=gray, fill=gray}, error bars/.cd, x dir=both, x explicit]
    coordinates { (-0.118,4) +- (0.072,0.078) };
\addplot+[morange, mark=square*, mark options={draw=morange, fill=morange}, error bars/.cd, x dir=both, x explicit]
    coordinates { (0.314,3) +- (0.204,0.176) };
\addplot+[mblue, mark=*, mark options={draw=mblue, fill=mblue},error bars/.cd, x dir=both, x explicit]
    coordinates { (0.322,2) +- (0.152,0.138) };
\addplot+[mgreen, mark=*, mark options={draw=mgreen, fill=mgreen}, error bars/.cd, x dir=both, x explicit]
    coordinates { (0.398,1) +- (0.168,0.152) };

\nextgroupplot[
    title={(b) ACC$_{\pm1}$ (\%)},
    xmin=60, xmax=95,
]
\addplot+[gray, mark=*,mark options={draw=gray, fill=gray}, error bars/.cd, x dir=both, x explicit]
    coordinates { (73.13,4) +- (5.22,5.23) };
\addplot+[morange, mark=square*, mark options={draw=morange, fill=morange}, error bars/.cd, x dir=both, x explicit]
    coordinates { (77.55,3) +- (8.16,8.16) };
\addplot+[mblue, mark=*, mark options={draw=mblue, fill=mblue},error bars/.cd, x dir=both, x explicit]
    coordinates { (76.12,2) +- (7.42,7.49) };
\addplot+[mgreen, mark=*, mark options={draw=mgreen, fill=mgreen}, error bars/.cd, x dir=both, x explicit]
    coordinates { (84.33,1) +- (6.33,5.67) };

\nextgroupplot[
    title={(c) MAE},
    xmin=0.55, xmax=1.25,
]
\addplot+[gray, mark=*, mark options={draw=gray, fill=gray}, error bars/.cd, x dir=both, x explicit]
    coordinates { (0.964,4) +- (0.064,0.056) };
\addplot+[morange, mark=square*, mark options={draw=morange, fill=morange}, error bars/.cd, x dir=both, x explicit]
    coordinates { (0.980,3) +- (0.170,0.190) };
\addplot+[mblue, mark=*, mark options={draw=mblue, fill=mblue},error bars/.cd, x dir=both, x explicit]
    coordinates { (0.866,2) +- (0.136,0.144) };
\addplot+[mgreen, mark=*, mark options={draw=mgreen, fill=mgreen}, error bars/.cd, x dir=both, x explicit]
    coordinates { (0.746,1) +- (0.136,0.154) };

\end{groupplot}
\end{tikzpicture}

\begin{tikzpicture} 
\begin{groupplot}[
    group style={
        group size=2 by 1,
        horizontal sep=2cm,
    },
    width=0.295\textwidth,
    height=3.6cm,
    ymin=0.3, ymax=4.7,
    ytick={1,2,3,4},
    yticklabels={X-LMC(ours), TICI-style, Inter-rater, Mean guess},
    yticklabel style={font=\fontsize{8pt}{\baselineskip}\selectfont},
    xticklabel style={font=\fontsize{8pt}{\baselineskip}\selectfont},
    y dir=normal,
    tick label style={font=\fontsize{8pt}{\baselineskip}\selectfont},
    grid=major,
    grid style={gray!15},
    axis line style={gray!60},
    error bars/x dir=both,
    error bars/x explicit,
    clip=false
]

\nextgroupplot[
    title={(d) Kappa},
    xmin=-0.08, xmax=0.68,
]
\addplot[draw=none, fill=yellow!12, forget plot] coordinates {
    (0.21,0.3) (0.40,0.3) (0.40,4.7) (0.21,4.7)} -- cycle;
\addplot[draw=none, fill=orange!10, forget plot] coordinates {
    (0.41,0.3) (0.60,0.3) (0.60,4.7) (0.41,4.7)} -- cycle;
\addplot[black!50, dashed, forget plot] coordinates {(0,0.3) (0,4.7)};
\addplot+[gray, mark=*,mark options={draw=gray, fill=gray}, error bars/.cd, x dir=both, x explicit]
    coordinates { (0.000,4) +- (0.000,0.000) };
\addplot+[morange, mark=square*,mark options={draw=morange, fill=morange}, error bars/.cd, x dir=both, x explicit]
    coordinates { (0.246,3) +- (0.186,0.184) };
\addplot+[mblue, mark=*, mark options={draw=mblue, fill=mblue}, error bars/.cd, x dir=both, x explicit]
    coordinates { (0.327,2) +- (0.167,0.153) };
\addplot+[mgreen, mark=*, mark options={draw=mgreen, fill=mgreen}, error bars/.cd, x dir=both, x explicit]
    coordinates { (0.430,1) +- (0.150,0.150) };

\nextgroupplot[
    title={(e) mF1},
    xmin=0.38, xmax=0.82,
]
\addplot+[gray, mark=*, mark options={draw=gray, fill=gray}, error bars/.cd, x dir=both, x explicit]
    coordinates { (0.439,4) +- (0.019,0.011) };
\addplot+[morange,mark options={draw=morange, fill=morange}, mark=square*, error bars/.cd, x dir=both, x explicit]
    coordinates { (0.612,3) +- (0.102,0.098) };
\addplot+[mblue, mark=*, mark options={draw=mblue, fill=mblue}, error bars/.cd, x dir=both, x explicit]
    coordinates { (0.663,2) +- (0.083,0.077) };
\addplot+[mgreen, mark=*, mark options={draw=mgreen, fill=mgreen}, error bars/.cd, x dir=both, x explicit]
    coordinates { (0.711,1) +- (0.081,0.079) };

\end{groupplot}
\end{tikzpicture}

\caption{
Multiclass (top) and binary (bottom) performance. Shaded $\kappa$ bands show fair [0.21 - 0.40] / moderate [0.41 - 0.60] agreement. Whiskers show 95\% confidence intervals.
}
\label{fig:agreement_forest_all}
\end{figure*}
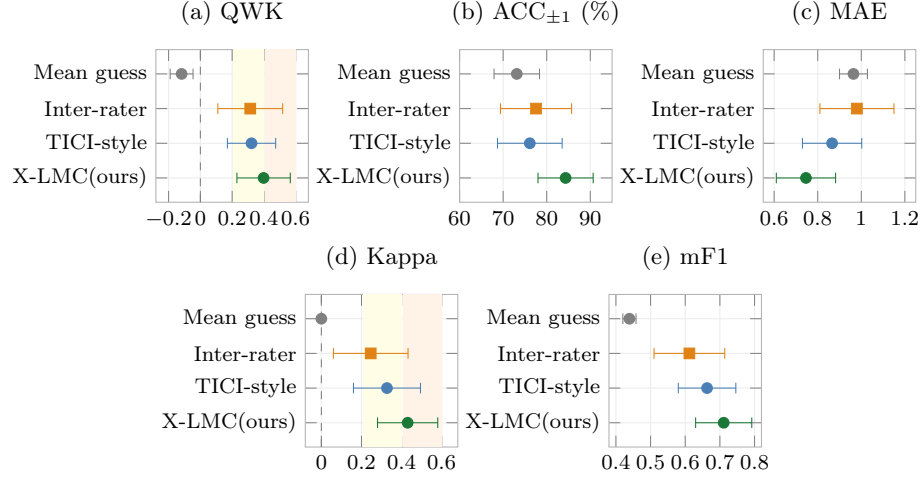

Comparative results across all evaluated metrics are summarized in Fig.~\ref{fig:agreement_forest_all}. In the multiclass task (Fig.~\ref{fig:agreement_forest_all}a--c), X-LMC achieves the highest agreement with reference labels across all models (QWK $=0.398$), with a higher point estimate than the adapted TICI-style baseline (QWK $=0.322$), while also yielding higher ACC$_{\pm1}$ ($84.33\%$ vs. $76.12\%$) and lower MAE ($0.746$ vs. $0.866$). Observed inter-rater agreement in this cohort was comparably modest (QWK $=0.314$), placing both the model and the human raters within the same fair-to-moderate agreement range; the model's point estimate is thus better interpreted as comparable to, rather than decisively exceeding, a single pair of expert raters, particularly given the overlapping confidence intervals at this cohort size (Fig.~\ref{fig:agreement_forest_all}a). In the binary task (Fig.~\ref{fig:agreement_forest_all}d--e), X-LMC achieves the highest Kappa ($0.430$) and mF1 $=0.711$ among evaluated methods, followed by the TICI-style baseline (Kappa=$0.327$, mF1 $=0.663$). Furthermore, all optimized deep learning models substantially outperform the mean-guess baseline across both ordinal and dichotomized settings. This performance gap demonstrates that the networks successfully capture task-specific hemodynamic representations rather than merely exploiting the structural target statistics of the cohort.

\begin{figure}[!b]
\centering

\begin{minipage}{0.8\textwidth}
\centering

\includegraphics[width=0.95\textwidth]{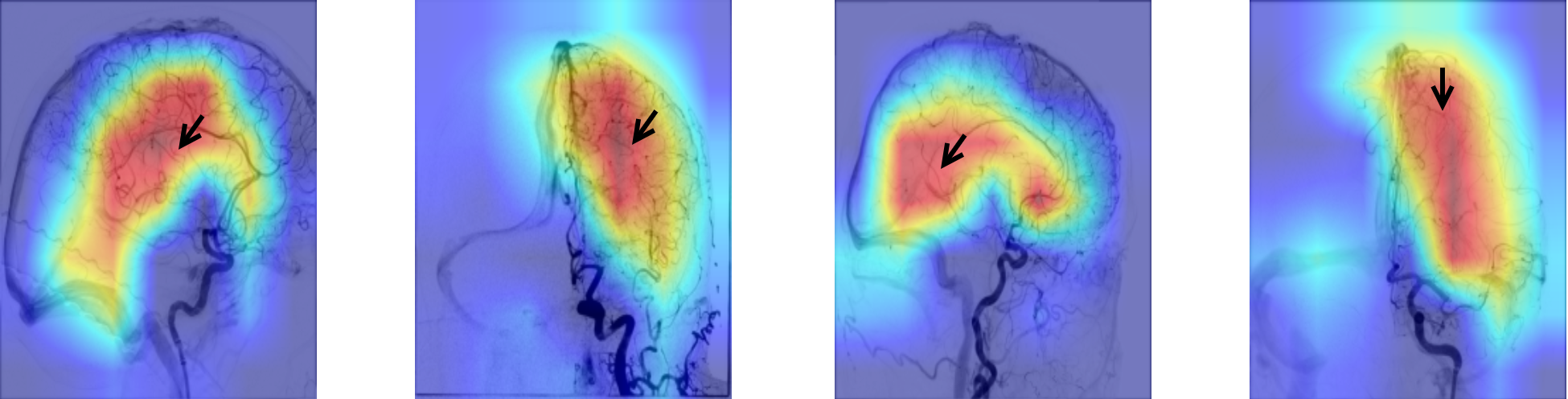}

\makebox[\linewidth]{%
\makebox[0.25\linewidth][c]{\fontsize{8pt}{\baselineskip}\selectfont (a)}%
\makebox[0.25\linewidth][c]{\fontsize{8pt}{\baselineskip}\selectfont (b)}%
\makebox[0.25\linewidth][c]{\fontsize{8pt}{\baselineskip}\selectfont (c)}%
\makebox[0.25\linewidth][c]{\fontsize{8pt}{\baselineskip}\selectfont (d)}%
}
\end{minipage}

\caption{Heatmap activations for good (a,b) and poor (c,d) ASITN/SIR scores. Arrows point to collaterals.}
\label{fig:gradcam}
\end{figure}

To qualitatively evaluate model behavior, frame-averaged Grad-CAM~\cite{selvaraju2020grad} activations were computed (Fig.~\ref{fig:gradcam}). Saliency maps for good collateral cases (Fig.~\ref{fig:gradcam}a--b) were frequently localized to distal cortical territories, consistent with the anatomical regions utilized during clinical ASITN/SIR assessment, while poor collateral cases (Fig.~\ref{fig:gradcam}c--d) showed relatively more central vascular activation, reflecting the absence of distal filling. These activation profiles indicate that the framework attends to physiologically relevant perfusion cues rather than relying on uninformative image regions.

\subsubsection{Ablations.} We conducted ablations following an 80/20 train/test split scheme (Table~\ref{tab:encoder_ablation}). We report QWK and Kappa as the primary agreement metrics for the ordinal and dichotomized tasks, respectively. Among the static spatial encoders, DINOv2-ViT-B achieved the best ordinal agreement (QWK = 0.364), outperforming EfficientNet-B0~\cite{tan2019efficientnet} and EfficientNetV2-S~\cite{tan2021efficientnetv2} despite the EfficientNet variants being fine-tuned rather than frozen, supporting the use of transformer-based representations for capturing fine-grained collateral features. Adding temporal modeling improved agreement across ablation variants. When keeping the fusion strategy fixed, replacing Uni-GRU with Bi-GRU increased QWK from 0.397 to 0.478 and Kappa from 0.250 to 0.429. When the temporal module was fixed to Bi-GRU, replacing concatenation with X-Attn yielded smaller but consistent relative gains in QWK and Kappa (8.0\% and 19.3\%, respectively), suggesting that cross-view attention provides an additional benefit beyond bidirectional temporal modeling alone.

\begin{table*}[t]
\centering
\caption{Ablation analysis in a 80/20 train/test scheme.}
\label{tab:encoder_ablation}
\begin{tabular}{l l c c c c}
\toprule
Ablation & Spatial Encoder & GRU & Fusion & QWK $\uparrow$ & Kappa $\uparrow$ \\
\midrule
\multirow{3}{*}{Spatial Encoder} & EfficientNet-B0   & --  & --     & 0.332 & 0.462 \\
                                  & EfficientNetV2-S  & --  & --     & 0.341 & 0.318 \\
                                  & DINOv2-ViT-B      & --  & --     & 0.364 & 0.456 \\
\midrule
\multirow{3}{*}{Temporal \& Fusion} & DINOv2-ViT-B & Uni & Concat & 0.397 & 0.250 \\
                                     & DINOv2-ViT-B & Bi  & Concat & 0.478 & 0.429 \\
                                     & DINOv2-ViT-B & Bi  & X-Attn & \textbf{0.516} & \textbf{0.512} \\
\bottomrule
\end{tabular}
\end{table*}

\section{Discussion \& Conclusion}

This work presents, to our knowledge, the first dedicated framework for automated ASITN/SIR collateral scoring from biplane DSA, establishing a foundation for exploring the feasibility and challenges of temporal deep learning for this task. Compared with spatiotemporal baselines adapted from related angiographic scoring tasks, X-LMC achieved higher performance across the evaluated metrics. The frozen DINOv2 backbone provided a stable, data-efficient representation for our limited cohort, while bidirectional X-Attn enabled synchronized AP and Lat projections to be jointly exploited across the angiographic sequence. The model’s agreement was comparable to inter-rater agreement, suggesting that performance is partly bounded by label ambiguity rather than architecture alone. Discrimination of adjacent ASITN/SIR grades, particularly the 2 vs.\ 3 boundary, proved most challenging, reflecting subtle differences in flow completeness and perfusion delay that are difficult to operationalize even for expert raters. The subjectivity inherent in the ASITN/SIR scale therefore represents a shared ceiling for both human and automated assessment, and adjudicated consensus annotation would be a meaningful step toward more reliable evaluation. This work should be read as an opening contribution mapping the feasibility and current boundaries of the task, rather than a mature clinical tool.

The clinical significance of automating this specific task extends beyond thrombectomy decision support. Because DSA provides direct, high-fidelity visualization of collaterals, DSA-derived scores offer a more reliable ground-truth reference~\cite{liu2018guidelines}. Automating this assessment at scale therefore unlocks retrospective cohort analysis that was previously infeasible due to the manual grading burden, enabling systematic association of collateral status with functional outcomes and imaging biomarkers, and paving the way toward cross-modal label propagation to pre-interventional modalities (e.g. CTA/CTP). Reliable collateral estimates also carry direct value for secondary treatment planning, neurorehabilitation scheduling, and hospital resource allocation, clinical insights that remain highly desired even after thrombectomy ~\cite{Campbell2013failure,binder2024leptomeningeal,liebeskind2025collaterals}.

To conclude, X-LMC demonstrates that automated architectures can reliably capture specialized microvascular kinetics. This work bridges the gap between raw biplane angiographic sequences and standardized collateralization indices, establishing a promising and reproducible foundation for automated hemodynamic profiling in stroke research.

\textit{Limitations.} Supervised training relied on a single expert reference label rather than adjudicated consensus. The cohort was further limited to occlusions of the M1 segment, reducing generalizability to other vascular territories. Future work should therefore prioritize consensus-labeled, multi-territory cohorts and external validation.

\begin{credits}
\subsubsection{\ackname} 
EdlR and BM are supported
by the Helmut Horten Foundation.

\subsubsection{\discintname}
The authors have no competing interests to declare that are relevant to the content of this article.
\end{credits}

%
%
%
\bibliographystyle{splncs04}
\bibliography{SWITCH_002}

\end{document}